# Concept-based Recommendations for Internet Advertisement

Dmitry I. Ignatov and Sergei O. Kuznetsov

Higher School of Economics, Department of Applied Mathematics
Kirpichnaya 33/5, Moscow 105679, Russia
{dignatov, skuznetsov}@hse.ru

**Abstract.** The problem of detecting terms that can be interesting to the advertiser is considered. If a company has already bought some advertising terms which describe certain services, it is reasonable to find out the terms bought by competing companies. A part of them can be recommended as future advertising terms to the company. The goal of this work is to propose better interpretable recommendations based on FCA and association rules.

## 1 Introduction

Contextual Internet advertising is a form of e-commerce. The largest revenues of the major players at this market, like search systems, are obtained from the so-called search sensitive advertisement, i.e, advertisement in a sense close to user queries. Here we consider the problem of detecting terms that can be interesting to an advertiser. Assume that a company $F$ has already bought some advertising terms which describe certain services. As a rule, there are already competing companies at the market, therefore it is reasonable to find terms bought by them. These terms can be compared to those bought by $F$ and part of them can be recommended as future advertising terms to $F$. The goal of this work is to propose well-interpretable recommendations based on FCA. The rest of the paper is organized as follows: First we recall main definitions from FCA and rule mining. Then we consider experimental data and the problem statement. Afterwards, we propose morphology-based and ontology-based metarules that can be derived without experimental data. We conclude the paper with experiments and their discussion.

## 2 Main definitions

First, we recall some basic notions from Formal Concept Analysis (FCA) [1]. Let $G$ and $M$ be sets, called the set of objects and attributes, respectively, and let $I$ be a relation $I \subseteq G \times M$: for $g \in G$, $m \in M$, $gIm$ holds iff the object $g$ has the attribute $m$. The triple $K = (G, M, I)$ is called a *(formal) context*. If $A \subseteq G$, $B \subseteq M$ are arbitrary subsets, then the *Galois connection* is given by the following *derivation operators*:

$$A' \stackrel{\text{def}}{=} \{m \in M \mid gIm \text{ for all } g \in A\},$$
$$B' \stackrel{\text{def}}{=} \{g \in G \mid gIm \text{ for all } m \in B\}.$$

If we have several contexts derivative operator of a context $(G, M, I)$ denoted by $(.)^I$.

The pair $(A, B)$, where $A \subseteq G$, $B \subseteq M$, $A' = B$, and $B' = A$ is called a *(formal) concept (of the context K)* with *extent* $A$ and *intent* $B$ (in this case we have also $A'' = A$ and $B'' = B$). For $B, D \subseteq M$ the *implication* $B \to D$ holds if $B' \subseteq D'$.

In data mining applications, an element of $M$ is called an *item* and a subset of $M$ is called an *itemset*.

The *support* of a subset of attributes (an itemset) $P \subseteq M$ is defined as $supp(P) = |P'|$. An itemset is *frequent* if its support is not less than a given *minimum support* (denoted by $min\_supp$). An itemset $P$ is closed if there exists no proper superset with the same support. The closure of an itemset $P$ (denoted by $P''$) is the largest superset of $P$ with the same support. The task of frequent itemset mining consists of generating all (closed) itemsets (with their supports) with supports greater than or equal to a specified $min\_supp$. An association rule is an expression of the form $I_1 \to I_2$, where $I_1$ and $I_2$ are arbitrary itemsets $(I_1, I_2 \subseteq A)$, $I_1 \cap I_2 = \emptyset$ and $I_2 \neq \emptyset$. The left side, $I_1$ is called *antecedent*, the right side, $I_2$ is called *consequent*. The support of an association rule $r : I_1 \to I_2$ [1] is defined as: $supp(r) = supp(I_1 \cup I_2)$. The *confidence* of an association rule $r$: $I_1 \to I_2$ is defined as the conditional probability that an object has itemset $I_2$, given that it has itemset $I_1$: $conf(r) = supp(I_1 \cup I_2)/supp(I_1)$. An association rule $r$ with $conf(r) = 100\%$ is an *exact* association rule (or implication [1]), otherwise it is an *approximate* association rule. An association rule is *valid* if $supp(r) \geq min\_supp$ and $conf(r) \geq min\_conf$. An itemset $P$ is a generator if it has no proper subset $Q(Q \subset P)$ with the same support. Let $FCI$ be the set of frequent closed itemsets and let $FG$ be the set of frequent generators. The *informative basis* for approximate association rules: $\mathcal{IB} = \{r : g \to (f \setminus g) | f \in FCI \land g \in FG \land g'' \subset f\}$.

## 3 Initial Data and Problem Statement

For experimentation we used data of US Overture [2], which were first transformed in the standard context form. In the resulting context $K = (G, M, I)$ objects from $G$ stay for advertising companies (advertisers) and attributes from $M$ stay for advertising terms (bids), $gIm$ means that advertiser $g$ bought term $m$. In the context $|G| = 2000$, $|M| = 3000$, $|I| = 92345$.

In our context, the number of attributes per object is bounded as follows: $13 \leq |g'| \leq 947$. For objects per attribute we have $18 \leq |m'| \leq 159$. From this context one had to compute formal concepts of the form (advertisers, bids) that represent market sectors. Formal concepts of this form can be further used for recommendation to the companies on the market, which did not buy bids contained in the intent of the concept. In other words, empty cell $(g, m)$ of the context can be considered as a recommendation to advertiser $g$ to buy bid $m$, if this advertiser bought other bids contained in the intent of any concept. This can also be represented as association rules of the form "If an advertiser bought bid $a$, then one can recommend this advertiser to buy term $b$" See [3] for the use of association rules in recommendation systems.

---
[1] In this paper we use absolute values, but the support of an association rule $r$ is also often defined as $supp(r) = supp(I_1 \cup I_2)/|O|$.

We consider the following context: $\mathbb{K}_{FT} = (F, T, I_{FT})$, where $F$ is the set of advertising firms (companies), $T$ is the set of advertising terms, or phrases, $fI_{FT}t$ means that firm $f \in F$ bought advertising term $t \in T$.

For constructing recommendations we used the following approaches and tools:

1. D-miner algorithm for detecting large market sectors as concepts;
2. Coron system for constructing association rules;
3. Construction of association metarules using morphological analysis;
4. Construction of association metarules using ontologies (thematic catalogs).

## 4 Standard approach to rule mining

### 4.1 Detecting large market sectors with D-miner.

D-miner is a freely available tool [4], [5] which constructs the set of concepts satisfying given constraints on sizes of extents and intents (icebergs and dual icebergs). D-miner takes as input a context and two parameters: minimal admissible extent and intent sizes and outputs a "band" of the concept lattice: all concepts satisfying constraints given by parameter values ($|intent| \geq m$ and $|extent| \geq n$, where $m, n \in \mathbb{N}$, see table 1).

**Table 1.** D-miner results.

| Minimal extent size | Minimal intent size | Number of concepts |
|---|---|---|
| 0 | 0 | 8 950 740 |
| 10 | 10 | 3 030 335 |
| 15 | 10 | 759 963 |
| 15 | 15 | 150 983 |
| 15 | 20 | 14 226 |
| 20 | 15 | 661 |
| 20 | 16 | 53 |
| 20 | 20 | 0 |

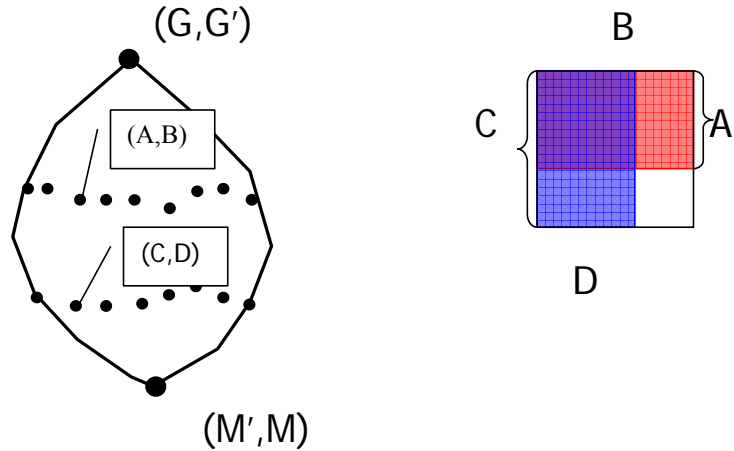

**Fig. 1.** A concept lattice and its band output by D-miner.

We give examples of intents of formal concepts for the case $|L| = 53$, where L is a concept lattice.

**Hosting market.**

{ affordable hosting web, business hosting web, cheap hosting, cheap hosting site web, cheap hosting web, company hosting web, cost hosting low web, discount hosting web, domain hosting, hosting internet, hosting page web, hosting service, hosting services web, hosting site web, hosting web }.

**Hotel market.**

{ angeles hotel los, atlanta hotel, baltimore hotel, dallas hotel, denver hotel, hotel chicago, diego hotel san, francisco hotel san, hotel houston, hotel miami, hotel new orleans, hotel new york, hotel orlando, hotel philadelphia, hotel seattle, hotel vancouver}

**Distance communication market.**

{ call distance long, calling distance long, calling distance long plan, carrier distance long, cheap distance long, company distance long, company distance long phone, discount distance long, distance long, cheap calling distance long, distance long phone, distance long phone rate, distance long plan, distance long provider, distance long rate, distance long service }

**Weight loss drug market.**

{ adipex buy, adipex online, adipex order, adipex prescription, buy didrex, buy ionamin, ionamin purchase, buy phentermine, didrex online, ionamin online, ionamin order, online order phentermine, online phentermine, order phentermine, phentermine prescription, phentermine purchase }

### 4.2 Recommendations based on association rules.

Using the Coron system (see [6]) we construct the informative basis of association rules [7]. We have chosen the informative basis, since it proposes a compact and effective way of representing the whole set of association rules. The results are given in table 2.

Table 2. Properties of informative basis.

| $min\_supp$ | $max\_supp$ | $min\_conf$ | $max\_conf$ | number of rules |
|---|---|---|---|---|
| 30 | 86 | 0,9 | 1 | 101 391 |
| 30 | 109 | 0,8 | 1 | 144 043 |

Here are some examples of association rules.

- $\{evitamin\} \rightarrow \{cvitamin\}$ supp=31 [1.55%]; conf=0.861 [86.11%]
- $\{gift\ graduation\} \rightarrow \{anniversary\ gift\}$, supp=41 [2.05%]; conf=0.820 [82.00%];

The value $supp = 31$ of the first rule means that 31 companies bought phrases "e vitamin" and "c vitamin". The value $conf = 0.861$ means that 86,1% companies that bought the phrase "e vitamin" also bought the phrase "c vitamin".

To make recommendations for each particular company one may use an approach proposed in [3]. For company $f$ we find all association rules, the antecedent of which contain all the phrases bought by the company, then we construct the set $T_u$ of unique advertising phrases not bought by the company $f$ before. Then we order these phrases by decreasing of confidence of the rules where the phrases occur in the consequences. If buying a phrase is predicted by several rules (i.e., the phrase is in the consequences of several rules), we take the largest confidence.

## 5 Mining metarules

### 5.1 Morphology-based Metarules

Each attribute of our context is either a word or a phrase. Obviously, synonymous phrases are related to same market sectors. The advertisers companies have usually thematic catalogs composed by experts, however due to the huge number of advertising terms manual composition of catalogs is a difficult task. Here we propose a morphological approach for detecting similar bids.

Let $t$ be an advertising phrase consisting of several words (here we disregard the word sequence): $t = \{w_1, w_2, \ldots, w_n\}$. A stem is the root or roots of a word, together with any derivational affixes, to which inflectional affixes are added [8]. The stem of word $w_i$ is denoted by $s_i = stem(w_i)$ and the set of stems of words of the phrase $t$ is denoted by $stem(t) = \bigcup_i stem(w_i)$, where $w_i \in t$. Consider the formal context $\mathbb{K}_{TS} = (T, S, I_{TS})$, where $T$ is the set of all phrases and $S$ is the set of all stems of phrases from $T$, i.e. $S = \bigcup_i stem(t_i)$. Then $tIs$ denotes that the set of stems of phrase $t$ contains $s$.

In this context we construct rules of the form $t \to s_i^{I_{TS}}$ for all $t \in T$, where $(.)^{I_{ts}}$ denotes the prime operator in the context $K_{TS}$. Then the a of the context $\mathbb{K}_{TS}$ (we call it a metarule, because it is not based on experimental data, but on implicit knowledge resided in natural language constructions) corresponds to $t \xrightarrow{FT} s_i^{I_{TS}}$, an association rule of the context $\mathbb{K}_{FT} = (F, T, I_{FT})$. If the values of support and confidence of this rule in context $\mathbb{K}_{FT}$ do not exceed certain thresholds, then the association rules constructed from the context $\mathbb{K}_{FT}$ are considered not very interesting.

**Table 3.** A toy example of context $\mathbb{K}_{FT}$ for "long distance calling" market.

| firm \ phrase | call distance long | calling distance long | calling distance long plan | carrier distance long | cheap distance long |
|---|---|---|---|---|---|
| $f_1$ | x |   | x |   | x |
| $f_2$ |   | x | x | x |   |
| $f_3$ |   |   |   | x | x |
| $f_4$ |   | x | x |   | x |
| $f_5$ | x | x |   | x | x |

**Table 4.** A toy example of context $\mathbb{K}_{TS}$ for "long distance calling" market.

| phrase \ stem | call | carrier | cheap | distanc | long | plan |
|---|---|---|---|---|---|---|
| call distance long | x |   |   | x | x |   |
| calling distance long | x |   |   | x | x |   |
| calling distance long plan | x |   |   | x | x | x |
| carrier distance long |   | x |   | x | x |   |
| cheap distance long |   |   | x | x | x |   |

Metarules of the following forms seem also to be reasonable. First, one can look for rules of the form $t \xrightarrow{FT} \bigcup_i s_i^{I_{TS}}$, i.e., rules, the consequent of which contain all terms containing at least one word with the stem common to a word in the antecedent term. Obviously, constructing rules of this type may result in the fusion of phrases related to different market sectors, e.g. "black jack" and "black coat". Second, we considered rules of the form $t \xrightarrow{FT} (\bigcup_i s_i)^{I_{TS}}$, i.e., rules with the consequent with the set of stems being the same as the set of stems of the antecedent. Third, we also propose to consider metarules of the form $t_1 \xrightarrow{FT} t_2$, where $t_2^{I_{TS}} \subseteq t_1^{I_{TS}}$. These are rules with the consequent being sets of stems that contain the set of stems of the antecedent.

**Example of metarules.**

- $t \xrightarrow{FT} s_i^{I_{TS}}$
  {last minute vacation} → {last minute travel}
  Supp= 19 Conf= 0,90
- $t \xrightarrow{FT} \bigcup_i s_i^{I_{TS}}$
  {mail order phentermine} → {adipex online order, adipex order, ...,
  phentermine prescription, phentermine purchase, phentermine sale}
  Supp= 19    Conf= 0,95
- $t \xrightarrow{FT} (\bigcup_i s_i)^{I_{TS}}$
  {distance long phone} → {call distance long phone, ...,
  carrier distance long phone, distance long phone rate, distance long phone service}
  Supp= 37    Conf= 0,88
- $t_1 \xrightarrow{FT} t_2, \quad t_2^{I_{TS}} \subseteq t_1^{I_{TS}}$
  {ink jet} → {ink}, Supp= 14    Conf= 0,7

## 5.2 Constructing ontologies and ontology-based metarules.

Here we use simple tree-like ontologies, where the closeness to the root of a tree defines generality of ontology concepts, which are advertisement phrases. For example, we use a manually constructed WordNet-like ontologies of market sectors. In our ontology of the pharmaceutical market the concept "pharmaceutical product" is more general than that of "vitamin." We introduce two operators acting on the set of advertising words $T$. *Generalization operator* $g_i(.) : T \to T$ takes a concept to a more general concept $i$ levels higher in the generality order. *Neighborhood operator* $n(.) : T \to T$ takes a concept to the set of sibling concepts.

Now we define two types of metarules for ontology: a generalization rule $t \to g_i(t)$ and a neighborhood rule $t \to n(t)$. These rules can also be considered as association rules of the context $\mathbb{K}_{FT} = (F, T, I_{FT})$, which allows one to understand which of them are good supported by data.
Examples of metarules for pharmaceutical market.
Rule of the form $t \to n(t)$, where $t =$ "$B\_VITAMIN$".

{$B\_VITAMIN$} → {$B\_COMPLEX\_VITAMIN, B12\_VITAMIN, C\_VITAMIN, \ldots$
$D\_VITAMIN, DISCOUNT\_VITAMIN, E\_VITAMIN, MINERAL\_VITAMIN, \ldots$
$MULTI\_VITAMIN, SUPPLEMENT\_VITAMIN, VITAMIN$}
Rules of the form $t \to g_1(t)$, where $t =$ "$B\_VITAMIN$", $g_1(t) =$ "$VITAMINS$".
{$B\_VITAMIN$} → {$VITAMINS$}.

## 6 Experimental Validation

For validation of association rules we used an adapted version of cross-validation. The training set was randomly divided into 10 parts, 9 of which were taken as the training set and the remaining part was used as a test set. By $A \xrightarrow{tr} B$ we denote an association rule generated on a training context. The confidence of this association rule measured on the test set, i.e.,

$$conf(A \xrightarrow{test} B) = \frac{|A^{I_{test}} \cap B^{I_{test}}|}{|A^{I_{test}}|}$$

shows the relative amount of companies that bought phrase $B$ having bought phrase $A$.

We constructed 10 sets of association rules for 10 different training sets 1800 companies each (with $min\_supp = 1,5\%$ and $min\_conf = 90\%$. The aggregated quality measure of the obtained rules is the average confidence:

$$average\_conf(Rules_i) = \frac{\sum\limits_{A \to B \in Rules} conf(A \xrightarrow{test} B)}{|Rules_i|},$$

where $Rules_i$ is the set of association rules obtained on the $i$-th training set. We also considered rules with $min\_conf \geq 0.5$ and computed averaged confidence, which was again averaged over 10 cases, $average\_conf = \frac{\sum\limits_{i=1}^{n} average\_conf(Rules_i)}{n}$.

Table 5. Results of cross-validation for association rules.

|       | Number of rules | Number of rules with sup > 0 | average_conf | Number of rules with min_conf=0.5 | average_conf (min_conf=0.5) |
|-------|-----------------|------------------------------|--------------|-----------------------------------|------------------------------|
| 1     | 147170          | 73025                        | 0,77         | 65556                             | 0,84                         |
| 2     | 69028           | 68709                        | 0,93         | 68495                             | 0,93                         |
| 3     | 89332           | 89245                        | 0,95         | 88952                             | 0,95                         |
| 4     | 107036          | 93078                        | 0,84         | 86144                             | 0,90                         |
| 5     | 152455          | 126275                       | 0,82         | 113008                            | 0,90                         |
| 6     | 117174          | 114314                       | 0,89         | 111739                            | 0,91                         |
| 7     | 131590          | 129826                       | 0,95         | 128951                            | 0,96                         |
| 8     | 134728          | 120987                       | 0,96         | 106155                            | 0,97                         |
| 9     | 101346          | 67873                        | 0,72         | 52715                             | 0,92                         |
| 10    | 108994          | 107790                       | 0,93         | 106155                            | 0,94                         |
| means | 115885          | 99112                        | 0,87         | 92787                             | 0,92                         |

The confidence of rules averaged over the test set is almost the same as the $min\_conf$ for the training set, i.e., $(0,9 - 0,87)/0,9 \approx 0,03$.

We used confidence measure also for validation of metarules. Support does not have much importance here, since we do not look for large markets or mostly sellable phrases, but stable dependencies of purchases. So, we considered only rules with confidence larger than 0.8 (or 0.9). Confidence and support for metarules are computed for the context $\mathbb{K}_{FT} = (F, T, I_{FT})$. We present the values of confidence and support in the tables for morphology-based metarules.

**Table 6.** Average support and average confidence for morphology-based metarules.

| Rule type | Average supp | Average conf | Number of rules |
|---|---|---|---|
| $t \xrightarrow{FT} s_i^{ITS}$ | 6 | 0,26 | 2389 |
| $t \xrightarrow{FT} \bigcup_i s_i^{ITS}$ | 6 | 0,24 | 456 |
| $t \xrightarrow{FT} (\bigcup_i s_i)^{ITS}$ | 12 | 0,40 | 1095 |
| $t \xrightarrow{FT} t_i$, such that $t_i^{ITS} \subseteq t^{ITS}$ | 15 | 0,49 | 7409 |
| $t \xrightarrow{FT} \bigcup_i t_i$, such that $t_i^{ITS} \subseteq t^{ITS}$ | 11 | 0,36 | 2006 |

We set the minimal support 0,5 and compute the number of rules of each group for which this threshold is exceeded. Table 5 shows that average_conf of these metarules is actually much higher (about 0,9).

**Table 7.** Average supp and conf for morphological metarules for $min\_conf = 0,5$.

| Rule types | Average supp | Average value of conf | Number of rules |
|---|---|---|---|
| $t \xrightarrow{FT} s_i^{ITS}$ | 15 | 0,64 | 454 |
| $t \xrightarrow{FT} \bigcup_i s_i^{ITS}$ | 15 | 0,63 | 75 |
| $t \xrightarrow{FT} (\bigcup_i s_i)^{ITS}$ | 18 | 0,67 | 393 |
| $t \xrightarrow{FT} t_i$ such that $t_i^{ITS} \subseteq t^{ITS}$ | 21 | 0,70 | 3922 |
| $t \xrightarrow{FT} \bigcup_i t_i$ such that $t_i^{ITS} \subseteq t^{ITS}$ | 20 | 0,69 | 673 |

From tables 6 and 7one can easily see that most confident and supported rules are of the form $t \xrightarrow{FT} \bigcup_i t_i$. Note that the use of morphology is completely automated and allows one to find highly plausible metarules without data on purchases. The rules with low support and confidence may be tested against recommendation systems such as Google AdWords, which uses the frequency of queries for synonyms. For validation of ontological rules we used Google service AdWords. 90% of recommendations (words) were contained in the list of synonyms output by AdWords.

# 7 Conclusion and further work

The obtained results show that a part of dependencies in databases for purchases of advertisement phrases may be detected automatically, with the use of standard means of computer linguistics. Along with methods of data mining, these approaches allows one to improve recommendations and propose good means of ranking, which is very important for making Top-N recommendations. Another advantage of the approach consists in the possibility of detecting related advertisement phrases not given directly in data. Results of FCA-based biclusterization show the possibility of detecting relatively large advertisement markets (with more than 20 participants) given by companies and advertising phrases. To improve the proposed approach we plan to use well-developed ontologies like WordNet for constructing ontology-based metarules.

# 8 Acknowledgements

This work was supported by the Scientific Foundation of Russian State University Higher School of Economics as a part of project 08-04-0022.

# References

1. Ganter, B., Wille, R.: Formal concept analysis: mathematical foundations. Springer, Berlin/Heidelberg (1999)
2. Zhukov, L.E.: Spectral clustering of large advertiser datasets. Technical report, Overture R&D (2004)
3. Sarwar, B.M., Karypis, G., Konstan, J.A., Riedl, J.: Analysis of recommendation algorithms for e-commerce. In: ACM Conference on Electronic Commerce. (2000) 158–167
4. Besson, J., Robardet, C., Boulicaut, J.F., Rome, S.: Constraint-based bi-set mining for biologically relevant pattern discovery in microarray data. Intelligent Data Analysis journal **9**(1) (2005) 59–82
5. Besson, J., Robardet, C., Boulicaut, J.F.: Constraint-based mining of formal concepts in transactional data. In Dai, H., Srikant, R., Zhang, C., eds.: PAKDD. Volume 3056 of Lecture Notes in Computer Science., Springer (2004) 615–624
6. Szathmary, L., Napoli, A.: CORON: A Framework for Levelwise Itemset Mining Algorithms. In: Suppl. Proc. of ICFCA '05, Lens, France. (2005) 110–113
7. Szathmary, L., Napoli, A., Kuznetsov, S.O.: ZART : A Multifunctional Itemset Mining Algorithm. Research Report 00001271, LORIA (2005)
8. David, C.: A dictionary of linguistics and phonetics. third edn. Oxford: Blackwell Publishers (1991)